\documentclass[conference]{ieeetran}  

\IEEEoverridecommandlockouts                              
\usepackage{graphicx} 
\usepackage{amsmath} 
\usepackage{amssymb}  
\usepackage{algorithm}
\usepackage{algorithmic}
\usepackage{xcolor}
\usepackage[font=scriptsize]{caption}

\usepackage[normalem]{ulem}

\usepackage{float}
\graphicspath{{figures/}}

\title{A Motion Taxonomy for Manipulation Embedding
}

\usepackage[font=scriptsize]{subfig}

\usepackage{amsmath}

\author{David Paulius, Nicholas Eales, and Yu Sun\\
\thanks{David Paulius, Nicholas Eales, and Yu Sun are with the Department of Computer Science and Engineering at the University of South Florida, Tampa, FL, USA. They are all members of the Robot Perception and Action Lab (RPAL). Nicholas was an undergraduate student when this work was done.
\newline(Contact email: \texttt{\{davidpaulius,yusun\}@usf.edu)}}
}

\graphicspath{{figures/}}
\usepackage{booktabs}

\begin{document}

\maketitle

\begin{abstract}
To represent motions from a mechanical point of view, this paper explores motion embedding using the motion taxonomy.
With this taxonomy, manipulations can be described and represented as binary strings called motion codes.
Motion codes capture mechanical properties, such as contact type and trajectory, that should be used to define suitable distance metrics between motions or loss functions for deep learning and reinforcement learning.
Motion codes can also be used to consolidate aliases or cluster motion types that share similar properties.
Using existing data sets as a reference, we discuss how motion codes can be created and assigned to actions that are commonly seen in activities of daily living based on intuition as well as real data.
Motion codes are compared to vectors from pre-trained Word2Vec models, and we show that motion codes maintain distances that closely match the reality of manipulation.
\end{abstract}

\IEEEpeerreviewmaketitle

\section{Introduction}


In robotics and AI, motion recognition is a crucial component to the understanding of the intent of humans and learning manipulations directly from demonstration.
In learning to recognize manipulations in activities of daily living (ADL), it is important to properly define motions or actions for use in classifiers.
However, it is very difficult to appropriately define or describe motions -- which we describe in human language in words -- in a way that is understood by robots.

Typically, motion recognition is achieved using classifiers, such as neural networks, to detect actions in sequences of events.
Networks are conventionally trained using one-hot vectors (for each motion class) for motion recognition; however, distances between motions -- i.e. distinguishing what makes one motion different to another -- are not innately measurable with such vectors.
Instead, \textit{word embedding} allows us to obtain a better vectorized representation of human language describing those actions, which can then be used to draw meaningful relationships in a high-dimensional space.
Essentially, words can be measured against one another for a variety of tasks, and they have been applied to affordance learning and grounding \cite{fang2018demo2vec,daruna2019robocse,roesler2019evaluation}.
One popular approach to learn embedding directly from natural language is Word2Vec \cite{mikolov2013efficient,mikolov2013distributed}.
However, the major drawback to Word2Vec, when applied in classifiers for motion recognition and analysis, is that vectors have no innate sense of mechanics or functionality required to execute those motions since they are trained on text.
A similar argument is made in other works that also consider embedding derived directly from manipulations over Word2Vec~\cite{daruna2019robocse}.
To elaborate further, with Word2Vec embedding, we cannot explain the difference between two types of actions, which may or may not share overlapping features, as distances between vectors are not functionally valid.
For instance, let us consider the labels `pour', `tap', and `poke'; when comparing them in pairs, with a pre-trained Word2Vec from Wikipedia \cite{yamada2018wikipedia2vec}, the labels `pour' and `tap' are closer to one another than `tap' and `poke', where the latter is considered to be mechanically closer.
Furthermore, Word2Vec embedding cannot capture multiple meanings of words.
In the prior example, the label `tap' can also refer to the noun for a source of water, which possibly explains why it is deemed more similar to `pour' than to `poke'.

Our objective in this paper is to introduce a suitable embedding of manipulations that considers mechanics of manipulations (from the point of view of the robot) for measuring distances.
Such a representation from the viewpoint of robots is important for robot learning and understanding~\cite{paulius2019survey}
With suitable distance metrics, motion classifiers can better discern between motion types or, in the event of uncertainty, suggest similar yet accurate labels for activity understanding.
We coined the term \textit{motion taxonomy} for this representation~\cite{paulius2019manipulation}.
With regards to existing taxonomies in robotics, grasp taxonomies have proved to be extremely useful in robotic grasp planning and in defining grasping strategies \cite{cutkosky1989grasp,worgotter2013simple,bullock2013hand,dai2013functional,feix2016grasp,abbasi2016grasp,marino2016data,nakamuracomplexities}. 
These studies further explored the dichotomy between power and precision grasps based on the way fingers secure objects contained within the hand.
However, there are no taxonomies that primarily focus on the mechanics of manipulation motions -- specifically contact and trajectory for each end-effector -- for defining motions and suitable metrics for motion types.
Such a taxonomy can also be used for consolidating motion aliases (i.e. words or expressions in human language) by representing them as binary codes, which may help to enforce grasp taxonomies for learning from demonstration.
This taxonomy can be used in a deep neural network that takes a manipulation sequence as input and outputs a representation of the motion type in the form of a binary-encoded string or code.
These codes can potentially be used for motion recognition, analysis, and generation.
In terms of robotic manipulation, using binary codes as word vectors for motions is better than word embedding from natural language like Word2Vec.
Furthermore, the taxonomy can be used in manipulation learning to identify skills that can be extended to other motions and to facilitate affordance learning similar to prior work~\cite{lopes2007affordance,ugur2014bootstrapping,moldovan2018relational,allevato2018}. 
In this paper, we 
show how this taxonomy can be used to define a representation that properly defines distances by comparing projections of various motions from the taxonomical embedding of labels to those from existing pre-trained Word2Vec models \cite{mikolov2013efficient,yamada2018wikipedia2vec,speer2017conceptnet}.

\section{Examining Motion Codes}
\label{sec:tax}
In this section, we describe the various attributes that are used to represent manipulations as motion codes using the motion taxonomy.
Briefly, the purpose of this taxonomy is to translate manipulations into a machine language for the purpose of motion recognition, analysis and generation.
Here, we define a manipulation motion to be any atomic action between \textit{active} and \textit{passive} objects; an active object is defined as a tool or utensil (or the combination of a robot's gripper or human hand and tool) that acts upon passive objects, which are objects that are acted upon as a result of motion.
Motions can be effectively embedded as vectors that relates them to motion feature space.
Motions labelled with motion codes avoid ambiguity from aliases for motion types, especially when translating between human languages.

\subsection{A Case for the Motion Taxonomy}
Deriving a representation of motions using the motion taxonomy was partially inspired by our own experiences with annotating labels for robot knowledge.
We have observed that among several annotators, inconsistency of labelling and defining motions was prevalent.
This happens especially with certain motion types that are hard to discern (such as deciding between the labels `cut', `slice' or `chop'), which requires revisiting all labels given to videos to ensure consistency.
Furthermore, this is also a problem encountered when using annotated data sets such as the MPII Cooking Activities Dataset \cite{MaxPlankIICooking} or EPIC-KITCHENS \cite{Damen2018EPICKITCHENS} since they may have their own labels that may not overlap with each other.
In some cases, labels can be very ambiguous and could be better described when adopting data sets for affordance learning
For instance, in EPIC-KITCHENS, one verb class provided is `adjust', which turns out to encompass several actions such as tapping, poking, pressing or rotating depending on types of switches; another example is the `insert' class, which encompasses actions such as pouring to picking-and-placing.

To potentially resolve these issues, we propose a representation scheme that deviates from natural language since an effective representation is important for robot learning~\cite{paulius2019survey}.
Binary-encoded strings called \textit{motion codes} will inherently define motions based on key traits defined in the taxonomy.
Ambiguity in human language labels or classes can be better handled if we represent them in an attribute space, especially if these can be automatically obtained from demonstration.
Different from this representation, neural networks that are used for motion recognition typically encode motion labels using one-hot vectors.
When training such networks, we use the cross entropy loss function, which is defined as:

\begin{equation*}
    L = -{\displaystyle\sum}_{k=1}^{N}{\mathbf{x}_{k}\log{\hat{\mathbf{x}}_{k}}}
\end{equation*}

\noindent where $N$ is the total number of classes, $\mathbf{x}_{k}$ is the ground-truth distribution, and $\mathbf{\hat{x}}_{k}$ is the predicted distribution.
For instance, if we have three labels `pour', `sprinkle', and `cut', these may be encoded with vectors $[1,0,0]$, $[0,1,0]$, and $[0,0,1]$ respectively; during the prediction stage, we can predict the label for a given manipulation sequence with the highest confidence using this equation.
Since cross entropy is used to determine how close predicted distributions are to the actual distribution using one-hot vectors, distances between classes would not matter since one-hot vectors are equidistant from one another. 
Although we can consider this as a distance metric between probabilities, this does not consider class features that can provide a better label for class instances.
Following the prior example, we do not get a sense of similarity between motions: pouring and sprinkling can be considered as closer motions than to cutting in terms of manipulation mechanics.

With Word2Vec embeddings \cite{mikolov2013efficient}, cosine distances between vectors suggest relatedness between word labels, where relatedness is determined by context.
These models are trained either using continuous bag-of-words (CBOW), n-grams or skip-grams to identify word pairs that are frequently used or seen together.
However, these vectors do not explicitly define why motions differ, which is one key purpose of motion codes; since Word2Vec derives vectors for singular words, we also can run into issues when defining variations of motions.
For example, pushing a solid or rigid object is mechanically different to pushing a soft object since the object we are pushing changes in shape, but we cannot represent these variations with Word2Vec.
With motion codes, we can be more descriptive with our characterization of motions.
It is important to note that the proposed motion taxonomy is not claimed to be the ideal way of representing motions; rather, it can be used to tentatively reduce the amount of features needed to label and compute meaningful distances between motions.

\subsection{Examining Characteristics of Motion Codes}
\begin{figure}[t]
	\centering
    \includegraphics[width=8cm,trim={0.9cm 0.75cm 0.5cm 0.65cm}, clip]{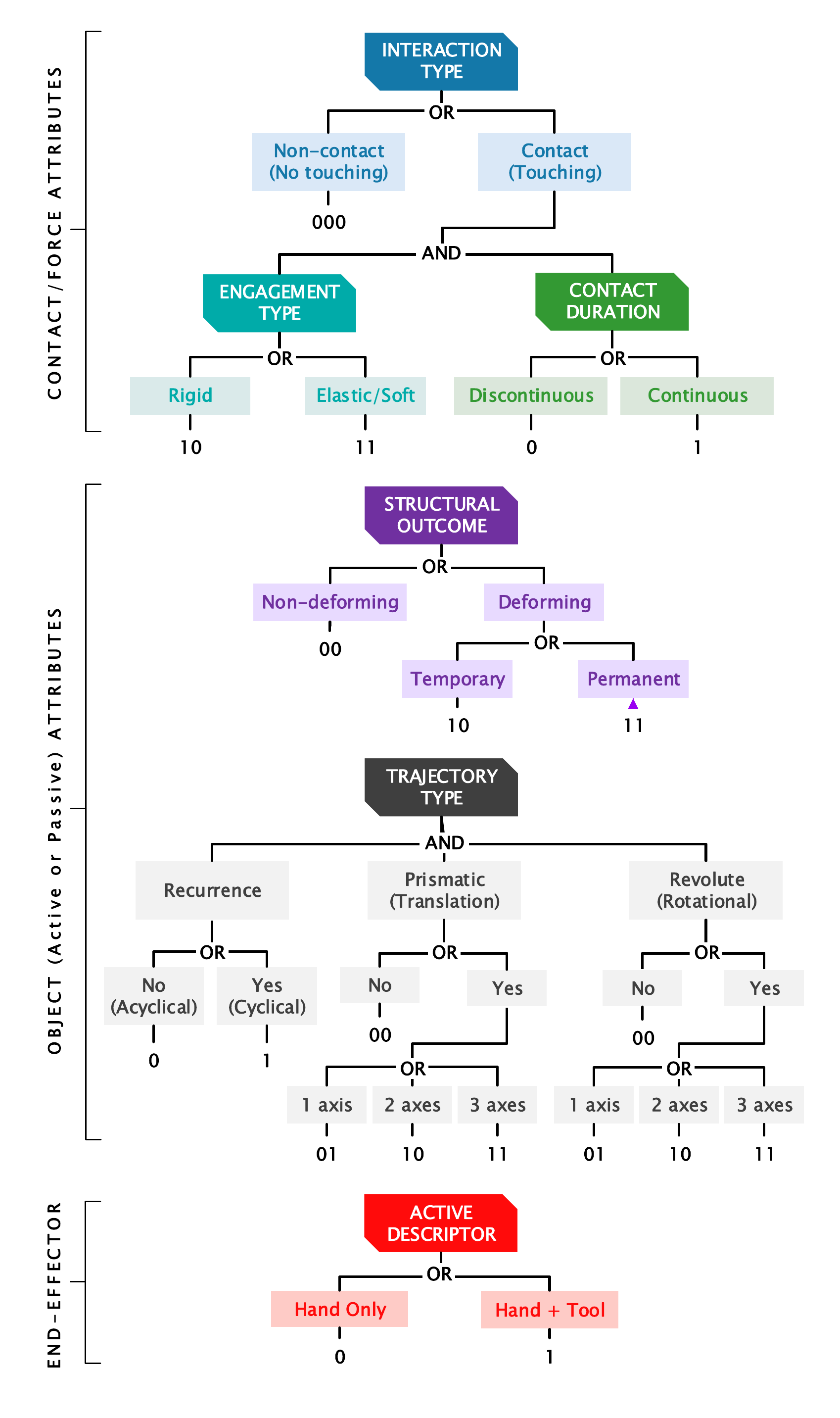}
    \caption{Illustration of the hierarchy of attributes in the motion taxonomy.  
    A motion code is formed by appending contact features, the active object's structural bits, the passive object's structural bits, the active trajectory and passive trajectory bits, and a bit descriptor of the active object by following the tree.
    }
    \label{fig:taxx}
\end{figure}

The mechanics of motions can be broken down into contact, force and trajectory.
Hence, our taxonomy considers the following attributes based on contact and trajectory information: contact interaction type, engagement type, contact duration, trajectory type and motion recurrence. 
Motion codes also indicate whether the active object is solely a hand/gripper or if it is a combination of a hand or tool.
Motion codes can be defined for each end-effector used in the manipulation.
When considering contact, we examine whether objects used in the manipulation make contact with one another and we describe what happens to these objects when this contact is established.
These features, shown in Figure \ref{fig:taxx}, are further described below.
\newline

\subsubsection{Describing Contact Type and Features}
Motion types can be classified as {\it contact} or {\it non-contact} interaction types.
Contact motion types are those that require contact between an active object (i.e. the actor's hands or the object that is typically grasped in the actor's hands) and passive object(s) (i.e. the object(s) that is/are manipulated upon when in contact with an active object) in the work space.
As opposed to the taxonomy found in \cite{paulius2019manipulation}, we may consider the hand or end-effector as a tool.
Conversely, non-contact motion types are those where no contact is established between active and passive objects or there is no force exerted upon passive objects.
Contact can be observed with vision (for instance, by the objects' borders or bounding boxes overlapping) or using force sensors mounted on objects.
An example of a contact motion is mixing, where the active tool makes contact with contents within a passive container.
As for a non-contact motion, pouring is a prime example: when pouring from one container to another, the active container held in the hand is not required to make contact with the passive receiving container.

Once an object interacts with another through physical contact, we classify their engagement as either {\it rigid} or {\it soft}.
Rigid engagement is where an active object's interaction with passive objects does not result in deformation -- i.e. their structure is not compromised or changed --, whereas soft engagement is where objects deform as a result of the interaction or the objects allow admittance or are permeable.
Furthermore, we can also consider the structural integrity (or state) of the objects used in order to describe deformation.
Active and passive objects can either undergo no deformation (non-deforming) or structural change (deforming).
We consider the cutting action as a soft engagement motion, as an active {\it knife} object will permanently deform the passive object into smaller pieces or units; even in the action of mixing items within a bowl, the contents within the bowl can be regarded as the passive objects being acted upon and deformed.
As for a rigid motion, actions such as tapping or poking a solid object show no structural change among objects as a result of contact.
In spreading with a knife, neither the knife nor the surface (like bread) incurs a significant change in their shape.
Deformation can be further distinguished as temporary or permanent, which is attributed to an object's material or texture.
For instance, when we squeeze a passive \textit{sponge} object, it returns to its original shape, signifying that this motion temporarily deforms it.
However, in the cutting example from before, this state change is permanent.
Poking or tapping an object could be classified as soft engagement if we engage an elastic object.

%

In addition to the prior attributes, it may be useful to note if the active tool has persistent contact with passive objects.
If the actor only makes contact for a short duration in the manipulation, we consider that contact to be \textit{discontinuous}; however, if the contact between the active tool and passive object is persistent, we consider that contact to be \textit{continuous}.
However, this perspective changes depending on what is considered to be the active object.
If we consider the robot's hand to be the active tool only, then we can assume that once it is grasping a tool for manipulation, there would be continuous contact between the hand and the tool.
This is why we consider the active object to be either the hand (if no tool acts upon other objects) or both the hand and tool as a unit (if there are other objects in the manipulation).
Contact duration can be determined visually (by timing the overlap of bounding boxes, for instance) or physically with force sensors.
For the sake of this work, we rely on demonstration videos to identify contact duration by intuition.
\newline

\subsubsection{Describing Trajectory of Motion}
We can describe an object's trajectory as {\it prismatic} (or translational), {\it revolute} (or rotational), or {\it both}.
Previously, we solely encoded the active object's observed trajectory in motion codes~\cite{paulius2019manipulation}, but in this work, we have included the passive object's trajectory.
Prismatic motions are manipulations where the object is moved along a certain axis or plane of translation.
Prismatic motions can be 1-dimensional (along a single axis), 2-dimensional (confined to a plane) or 3-dimensional (confined to a manifold space); this can be interpreted as having 1 to 3 DOF of translation.
Revolute motions, on the other hand, are manipulations where the object is rotated about an axis or plane of rotation; a robot performing such motions would rely on revolute joints to execute manipulations of this nature.
Similar to prismatic motions, revolute motions can also range from 1-dimensional to 3-dimensional motion (i.e. from 1 to 3 DOF of rotation); typically, revolute motions are confined to a single axis of rotation in world space.
A motion is not limited to one trajectory type, as these properties are not mutually exclusive; therefore, we can say that a motion can be prismatic-only, revolute-only, neither prismatic nor revolute or both prismatic and revolute.
From the perspective of the active object, an example of a prismatic-only manipulation is chopping with a knife since the knife's orientation is usually fixed, while an example of a revolute-only motion is fastening a screw into a surface using a screwdriver.
However, a motion such as scooping with a spoon will usually require both prismatic and revolute movements to complete the action.

We can also describe a motion's trajectory by its \textit{recurrence}, which describes whether the motion exhibits repetitive behaviour in the tool's movement. 
A motion can be \textit{acyclical} or \textit{cyclical}, which may be useful depending on the context of motion.
For instance, mixing ingredients in a bowl may be repeated until the ingredients have fully blended together, or in the case of loosening a screw, the screwdriver will be rotated until the screw is completely out of the surface.
Learned acyclical motions can be made cyclical simply by repeating them, which is a decision that can be left up to the robot during motion generation if it is not finished with its task or it failed to execute the manipulation successfully.


\begin{table}[t]
\centering
\caption{Motion codes for manipulations based on the taxonomy 
illustrated in Figure \ref{fig:taxx}.
    The attributes of each motion correspond to those observed in source demonstrations.
    These codes are best viewed in colour, as each binary bit is colour-coded based on Figure \ref{fig:taxx}.
    Motion codes are 17 bits long.
    Underlined substrings correspond to the active object.
}
\label{tab:motion_code}
\begin{tabular}{p{2.3cm}l}
\toprule[1pt]
\textit{\textbf{Motion Code}} & \textit{\textbf{Motion Types}} \\ 
\midrule[1pt]

\textbf{{\color{teal}000}{\color{violet}\underline{00}{00}}\underline{00001}{00000}{\color{red}1}}	&	pour \\

\textbf{{\color{teal}000}{\color{violet}\underline{00}{00}}\underline{10100}{00000}{\color{red}1}}	&	sprinkle \\

\textbf{{\color{teal}100}{\color{violet}\underline{00}{00}}\underline{00100}{00000}{\color{red}0}}	&	poke, press (button), tap \\

\textbf{{\color{teal}101}{\color{violet}\underline{00}{00}}\underline{00000}{00000}{\color{red}0}}	&	grasp, hold \\

\textbf{{\color{teal}101}{\color{violet}\underline{00}{00}}\underline{00001}{00001}{\color{red}0}}	&	open/close (jar), rotate, turn (key, knob), twist \\

\textbf{{\color{teal}101}{\color{violet}\underline{00}{00}}\underline{00100}{00000}{\color{red}1}}	&	spread, wipe \\

\textbf{{\color{teal}101}{\color{violet}\underline{00}{00}}\underline{00100}{00100}{\color{red}0}}	&	move, push (rigid) \\


\textbf{{\color{teal}101}{\color{violet}\underline{00}{00}}\underline{00101}{00101}{\color{red}0}}	&	flip (hand) \\

\textbf{{\color{teal}101}{\color{violet}\underline{00}{00}}\underline{00101}{00101}{\color{red}1}}	&	flip (turner, spatula) \\

\textbf{{\color{teal}101}{\color{violet}\underline{00}{00}}\underline{01000}{00000}{\color{red}1}}	&	spread, wipe (surface) \\

\textbf{{\color{teal}101}{\color{violet}\underline{00}{00}}\underline{01000}{00001}{\color{red}0}}	&	open/close (door) \\

\textbf{{\color{teal}101}{\color{violet}\underline{00}{00}}\underline{01000}{01000}{\color{red}0}}	&	move (2D), insert (placing), pick-and-place \\


\textbf{{\color{teal}101}{\color{violet}\underline{00}{00}}\underline{10001}{10001}{\color{red}1}}	&	fasten, loosen (screw) \\

\textbf{{\color{teal}101}{\color{violet}\underline{00}{00}}\underline{10001}{10001}{\color{red}0}}	&	shake (revolute) \\

\textbf{{\color{teal}101}{\color{violet}\underline{00}{00}}\underline{10100}{10100}{\color{red}0}}	&	shake (prismatic) \\


\textbf{{\color{teal}110}{\color{violet}\underline{00}{10}}\underline{00100}{00000}{\color{red}1}}	&	dip \\

\textbf{{\color{teal}110}{\color{violet}\underline{00}{10}}\underline{00101}{00100}{\color{red}1}}	&	scoop (liquid) \\

\textbf{{\color{teal}110}{\color{violet}\underline{00}{11}}\underline{00101}{00100}{\color{red}1}}	&	scoop \\

\textbf{{\color{teal}110}{\color{violet}\underline{11}{00}}\underline{00100}{00000}{\color{red}1}}	&	crack (egg) \\

\textbf{{\color{teal}111}{\color{violet}\underline{00}{00}}\underline{00100}{00000}{\color{red}1}}	&	insert, pierce \\

\textbf{{\color{teal}111}{\color{violet}\underline{00}{10}}\underline{00000}{00000}{\color{red}0}}	&	squeeze (in hand, elastic) \\

\textbf{{\color{teal}111}{\color{violet}\underline{00}{10}}\underline{00101}{00101}{\color{red}0}}	&	fold, unwrap, wrap \\

\textbf{{\color{teal}111}{\color{violet}\underline{00}{10}}\underline{11000}{00000}{\color{red}1}}	&	beat, mix, stir (liquid) \\

\textbf{{\color{teal}111}{\color{violet}\underline{00}{11}}\underline{00000}{00000}{\color{red}0}}	&	squeeze (in hand) \\




\textbf{{\color{teal}111}{\color{violet}\underline{00}{11}}\underline{00100}{00100}{\color{red}0}}	&	flatten, press, squeeze, pull apart, peel (hand) \\

\textbf{{\color{teal}111}{\color{violet}\underline{00}{11}}\underline{00100}{00000}{\color{red}1}}	&	chop, cut, mash, peel, scrape, shave, slice \\





\textbf{{\color{teal}111}{\color{violet}\underline{00}{11}}\underline{00100}{10001}{\color{red}0}}	&	roll \\


\textbf{{\color{teal}111}{\color{violet}\underline{00}{11}}\underline{01000}{00000}{\color{red}1}}	&	saw, cut (2D), slice (2D) \\

\textbf{{\color{teal}111}{\color{violet}\underline{00}{11}}\underline{11000}{00000}{\color{red}1}}	&	beat, mix, stir \\

\textbf{{\color{teal}111}{\color{violet}\underline{10}{00}}\underline{00100}{00100}{\color{red}1}}	&	brush, sweep, spread (brush) \\

\textbf{{\color{teal}111}{\color{violet}\underline{10}{00}}\underline{01000}{01000}{\color{red}1}}	&	brush, sweep (surface) \\

\textbf{{\color{teal}111}{\color{violet}\underline{11}{00}}\underline{00000}{00100}{\color{red}1}}	&	grate \\

\bottomrule[1pt]
\end{tabular}
\end{table}

\subsection{Translating Motions to Code}
\label{sec:foon}
We now discuss how motion codes can be assigned to motions using the example of the cutting action.
Using the flowchart shown as Figure \ref{fig:taxx}, we construct codes in the following manner: first, we ascertain whether the motion is contact or non-contact.
In cutting, the active knife object makes contact with the passive object, and so we will follow the contact branch.
If the motion was better described as non-contact, then we will start with the string \textit{`000'}.
Since there is contact, we then describe the type of engagement between the objects and how long the contact duration is throughout the action.
Following our example, the knife cuts through an object and maintains contact with it for the entirety of the manipulation, hence making it a soft engagement (\textit{`11'}) and with continuous contact (\textit{`1'}).
After describing contact, we describe the state of the active and passive objects after the manipulation occurs.
In our example, the active object does not deform (\textit{`00'}) while the passive object deforms permanently since the knife cuts it into a different state  (\textit{`11'}).
After describing the structural integrity of the objects, we then describe their trajectories.
When cutting an object, the active trajectory is typically a 1D prismatic motion as we swing the knife up and down and without any rotation (\textit{`00100'}), while there is no passive trajectory (\textit{`00000'}), as the passive object is usually immobile.
If we are observing repetition in cutting, then we would assign the recurrent bit \textit{`1'} instead of \textit{`0'} in the active trajectory substring.
Finally, we indicate whether the active object is solely the hand or hand/tool pair; 
in our example, we would assign it a bit of \textit{`1'} since we have a hand and knife pairing as an active object.
With all of these substrings, we end up with the single motion code \textit{`11100110010000001'}.

We compiled a list of motion labels that can be found across several sources of manipulation data such as EPIC-KITCHENS, MPII Cooking Activities, FOON \cite{Paulius2016,Paulius2018} (inspired by our prior work on object affordance for robot manipulation \cite{Ren2013,lin2015robot}), and Daily Interactive Manipulations (DIM) \cite{huang2019dataset}, and we show their respective codes in Table \ref{tab:motion_code}.
Several motions can share the same motion code due to common mechanics, such as cutting and peeling since they are both 1D-prismatic motions that permanently deform the passive objects.
We can also account for variations in manipulation; for instance, certain motions like mixing and stirring can either temporarily deform or permanently deform the target passive object, which depends on its state of matter, or we can identify non-recurrent or recurrent variants of motion.
It is important to note that motion codes can be assigned to each hand or end-effector used in a manipulation since they are not necessary to perform the same manipulation in the same action.
For instance, when chopping items, usually it is necessary to hold the object in place with one hand and then use the knife to chop with the other.
Because of this, the structural or state outcome of performing those actions could be extrinsic to the actions; in the aforementioned example, the passive object deforms but it is not directly an outcome of just holding the object.
In Table \ref{tab:motion_code}, we simplify this to the single-handed perspective of performing those actions.

\subsection{Obtaining Motion Codes from Demonstration}
Ideally, a neural network (or a collection of networks for a group of attributes) can be developed to output codes for different motion types.
In detail, such a network structure would assign motion codes to a series of segmented actions as seen in demonstration videos; rather than learning to detect and output a single motion code, an ensemble of classifiers that can separately identify parts of the hierarchy in Figure \ref{fig:taxx} can be used to build substrings that could then be appended together as a single, representative string.
As a result of such a network structure, one could also obtain motion features that may facilitate motion recognition tasks.

Representing manipulations in an attribute space as motion codes can be likened to the idea behind zero-shot learning (ZSL); just as in ZSL, even if certain class instances are not known, motion code vectors can be used as a guide to assign codes to unknown manipulations and to possibly learn new actions, granted that we know how to execute similar actions.

\section{Evaluation of the Taxonomy}
Having understood the taxonomy and identified motion codes for manipulations in ADL, we demonstrate how suitable they are for representing motion labels.
In particular, we focus on how motion codes can produce embeddings whose distances are meaningful based on their attributes.
Our evaluation is done in two parts: first, we show how the motion code assignment corresponds to actual data.
Second, we contrast motion codes to the unsupervised word embedding method Word2Vec \cite{mikolov2013efficient,mikolov2013distributed}, which learns vectorized representations of words directly from natural language, to show that it is not suitable to derive accurate motion embedding.
We used pre-trained models trained on Google News \cite{mikolov2013distributed,mikolov2013efficient}, Wikipedia~\cite{yamada2018wikipedia2vec}, and Concept-Net~\cite{speer2017conceptnet}; although these are not solely trained with knowledge sources of manipulations nor physical properties, these models are commonly used for deep learning tasks in robotics, AI, and computer vision.

\subsection{Support for Motion Codes}

\begin{figure}[t]
	\centering
	\includegraphics[width=5cm]{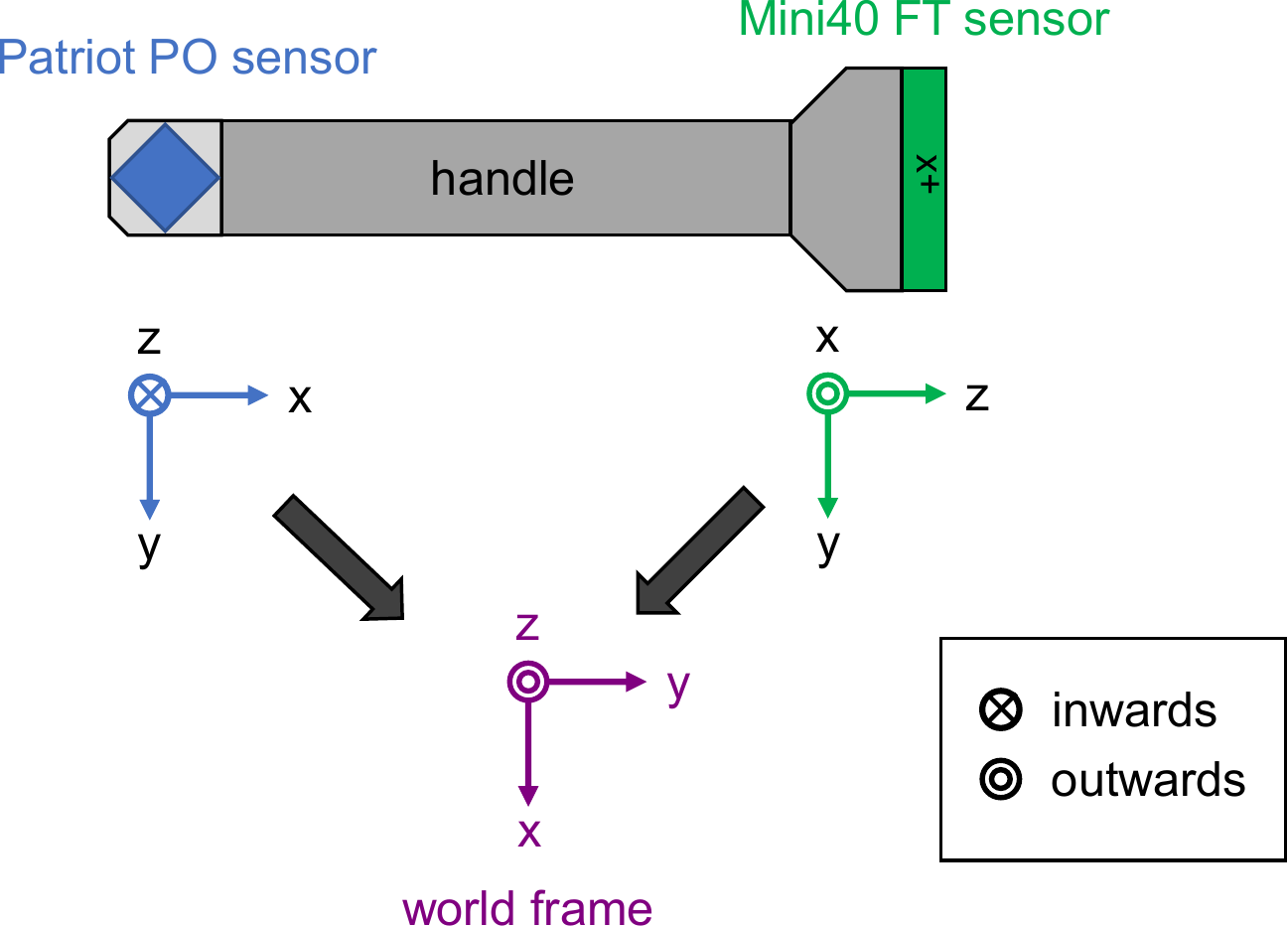}
	\caption{An illustration of adapter used for data collection in \cite{huang2019dataset} (best viewed in colour). The Patriot sensor (depicted in blue) records position and orientation, while the ATI Mini40 sensor (depicted in green) records force and torque.
	They are aligned to the world frame (depicted in purple) for analysis.
}
	\label{fig:adapter}
\end{figure}

\begin{figure}[t]
    \centering
    \subfloat[Projection of trajectory via PCA (stirring)] {{\includegraphics[width=7cm,trim={3.5cm 9cm 4cm 9.3cm},clip]{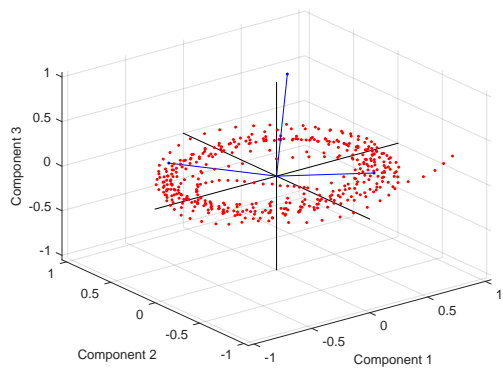}}
    \label{fig:PC-a}}
    
    \subfloat[Cosine similarity: linear velocity vs. PCs] {{\includegraphics[width=7cm,trim={4.3cm 9.2cm 4.3cm 9.7cm},clip]{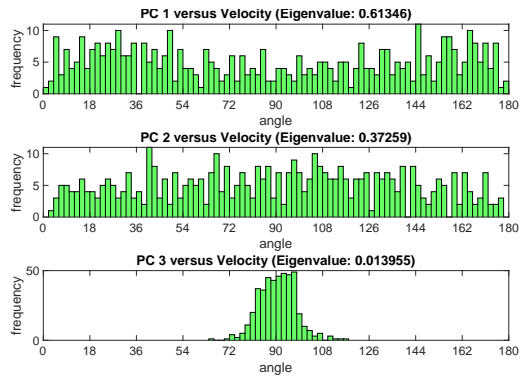}}
    \label{fig:PC-b}}    
    \caption{Example of how PCA can be applied to recorded position data to derive prismatic bits for the `stir' motion.
    In Figure \ref{fig:PC-a}, the trajectory's points lie on a plane, thus suggesting that this is a 2D prismatic motion.
    In Figure \ref{fig:PC-b}, which shows a histogram of the number of velocity vectors and their similarity to each PC, it further supports that the motion primarily lies in PCs 1 and 2 (capturing $\sim$99\% of variance).
    It can also be observed from the projection that this trajectory is recurrent since the motion is cyclical.}
    \label{fig:PCA}
\end{figure}

\begin{figure}[t]
    \centering
    \subfloat[Axis $K$ vs. tool's principal axis $y$ (loosen screw) (X-Y view)] {{\includegraphics[width=6.7cm,trim={0.5cm 0.3cm 1.2cm 0.7cm},clip]{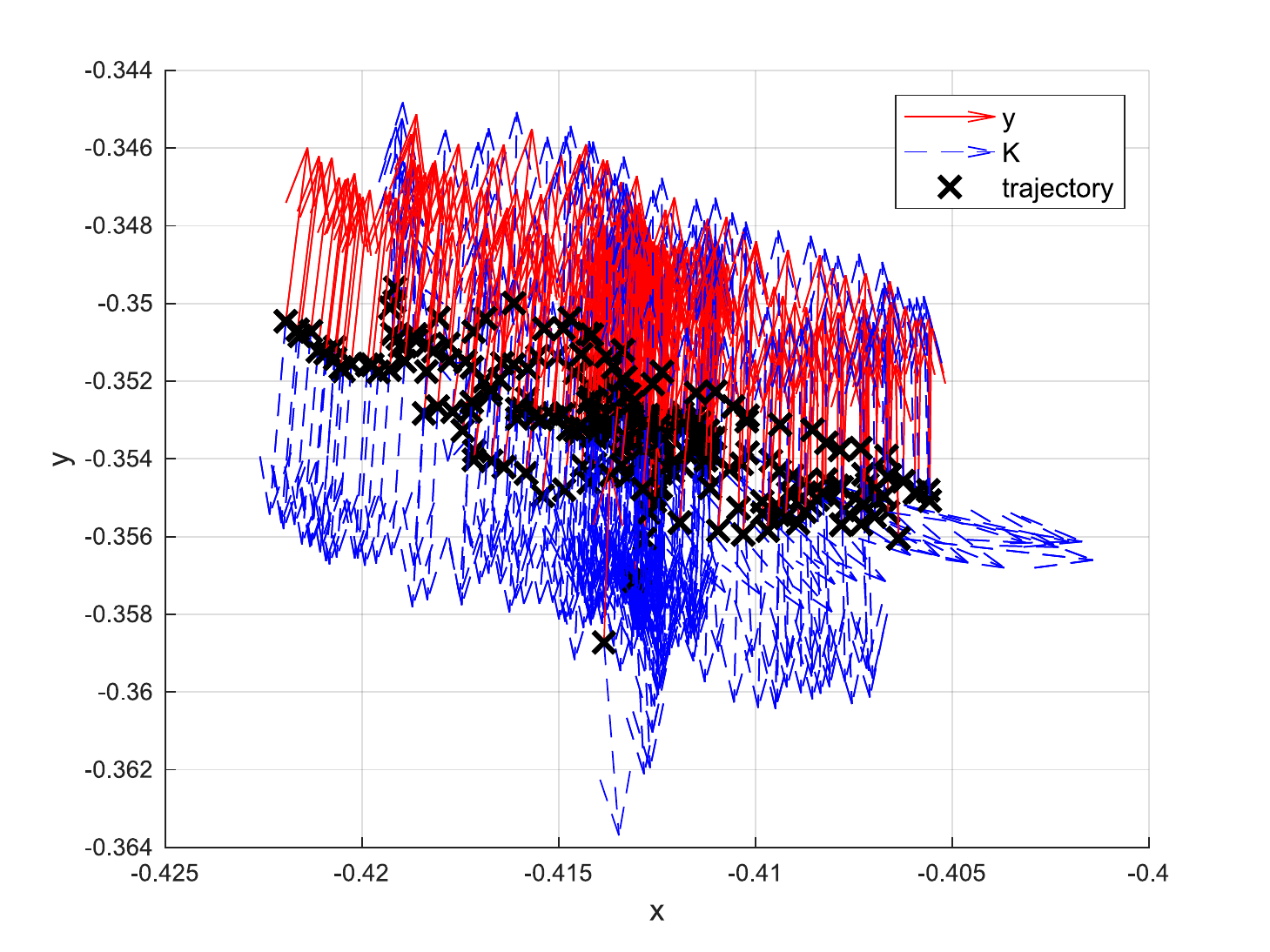}}
    \label{fig:ROT-a}} 

    \subfloat[Cosine similarity: axis $K$ vs. tool's principal axes] {{\includegraphics[width=6.7cm,trim={4.2cm 9cm 4.4cm 9.85cm},clip]{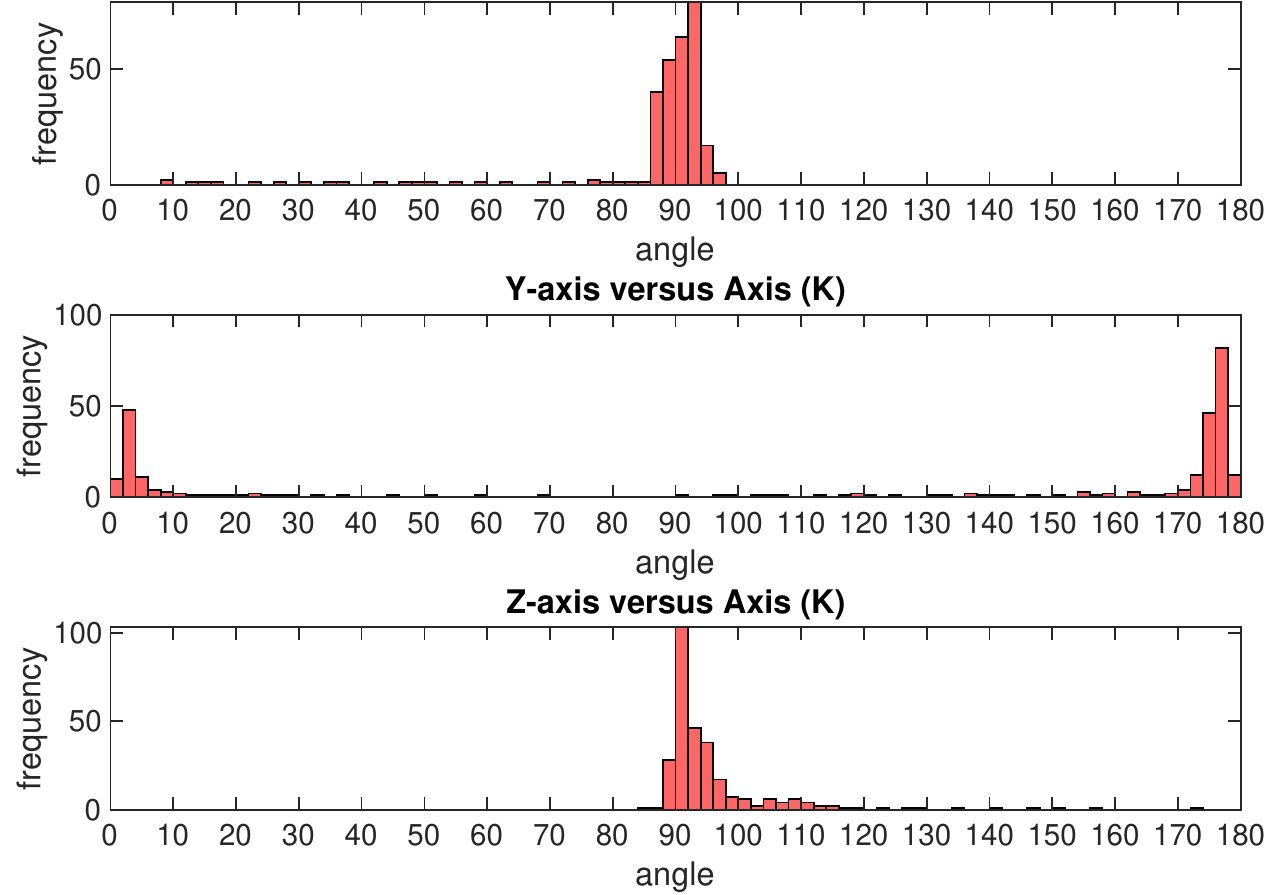}}
    \label{fig:ROT-b}}
    
    \subfloat[Degree of rotation about all principal axes] {{\includegraphics[width=6.7cm,trim={4.2cm 9cm 4.5cm 9.85cm},clip]{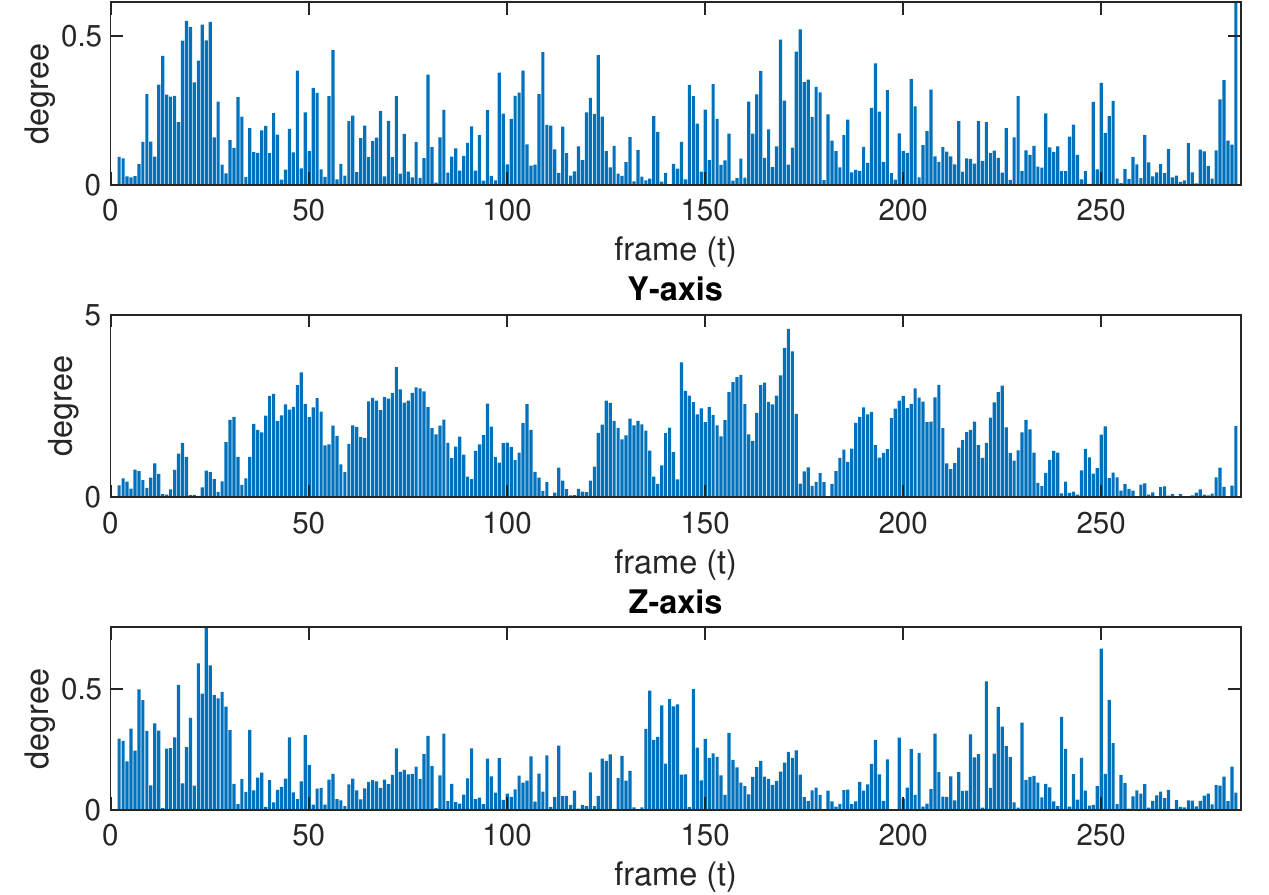}}
    \label{fig:ROT-c}}    
    \caption{Example of how the axis-angle representation can be used to identify revolute properties for the `loosen screw' motion (best viewed in colour).
    In Figure \ref{fig:ROT-a}, based on Figure \ref{fig:adapter}, we show how the axis vector $K$ (in blue), which is obtained from rotation matrices~\cite{spong2020robot}, aligns with the tool's principal axis $y$ (in red) at each trajectory point. 
    This is further supported by Figure \ref{fig:ROT-b}, which compares each frame's axis $K$ to the tool's principal axes with cosine similarity. 
    In Figure \ref{fig:ROT-c}, we graph the change in rotation about each axis with respect to the last frame's orientation.
    Figures \ref{fig:ROT-b} and \ref{fig:ROT-c} suggest rotation about the y-axis, hence making it a 1D revolute motion.}
    \label{fig:ROT}
\end{figure}

Preferably, motion codes are derived directly from demonstration data.
Several modalities of data such as trajectory, force, and vision can be used to determine the attributes that best describe said manipulations.
Using provided position and orientation data, which is available in data sets such as DIM \cite{huang2019dataset}, we can ascertain the trajectory type for several motions in which there is an active tool or object being manipulated.

To determine the prismatic trajectory type, we can use methods such as principal component analysis (PCA) to find the number of axes (which would be transformed into principal components, or PCs) that captures the most variance of the trajectory.
PCA has conventionally been used for feature extraction and dimension reduction, where one can obtain a subset of features (i.e. DOF) that will sufficiently capture data.
Here, we considered that the number of DOF for a motion is reflected by the number of PCs that would capture about 90\% of variance.
Motions such as flipping with a turner are effectively 1D (and in minor cases 2D) motions because a single PC captures about 90\% of the variance of those trials.
Mixing, beating and stirring (which are all variations of the same motion) data confirm that the motion is 2D since both the 1st and 2nd PCs met our requirements; this can be observed in the projection shown as Figure \ref{fig:PCA}.
One can compare the derived PCs to the velocity (i.e. directional vectors between trajectory frames) to also clarify whether or not motions exist within those dimensions using cosine similarity.
Should the velocity vectors align with the PCs, we would expect values closer to 0$^{\circ}$ or 180$^{\circ}$.
In Figure \ref{fig:PC-b}, not only does the 3rd PC contribute very little to capture the motion, but it is normal to velocity (since the histogram shows a prevalence of vectors with cosine similarity peaking around 90$^{\circ}$).

To determine the revolute trajectory type, we can convert the position and orientation data to rotation matrices and measure the amount of rotation about the principal axis of the active tool.
The axis-angle representation (which represents a frame as a vector $K$ and an angle of rotation $\theta$) derived from rotation matrices can also be used to compute the angle of rotation based on $\theta$.
A significant rotation about this principal axis suggests that there is at least one axis of rotation.
In Figure \ref{fig:ROT}, we illustrate how we can extract revolute properties for the motion of loosening a screw.
Given the tool's principal axes are defined as in Figure \ref{fig:adapter}, we expect that the operation of a screwdriver would require major rotation about the y-axis.
In Figure \ref{fig:ROT-a}, one can see that the axis vector $K$ is predominantly pointing in the opposite direction of the tool's axis (which is also supported by Figure \ref{fig:ROT-b}, which shows that the cosine similarity values peak at 0$^{\circ}$ or 180$^{\circ}$), suggesting that there is anti-clockwise (or negative) rotation.
Rotation about this axis is further supported by Figures \ref{fig:ROT-b} and \ref{fig:ROT-c}.

\begin{figure*}[h!]
	\centering
    \subfloat[Motion Codes (Contact)]{\includegraphics[width=8.5cm]{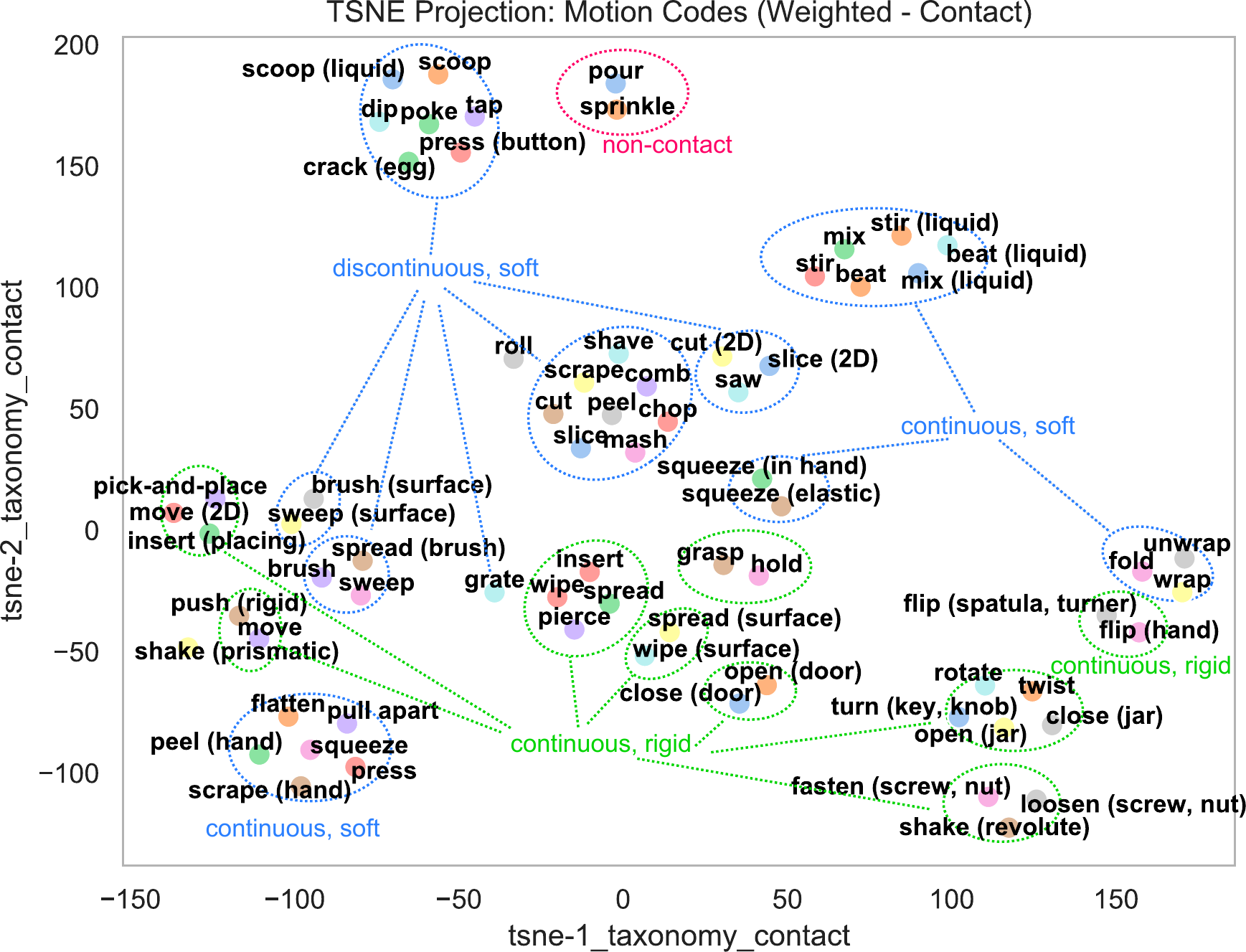}
    \label{fig:embed-mot_c}}
    \subfloat[Motion Codes (Trajectory)]{\includegraphics[width=8.5cm]{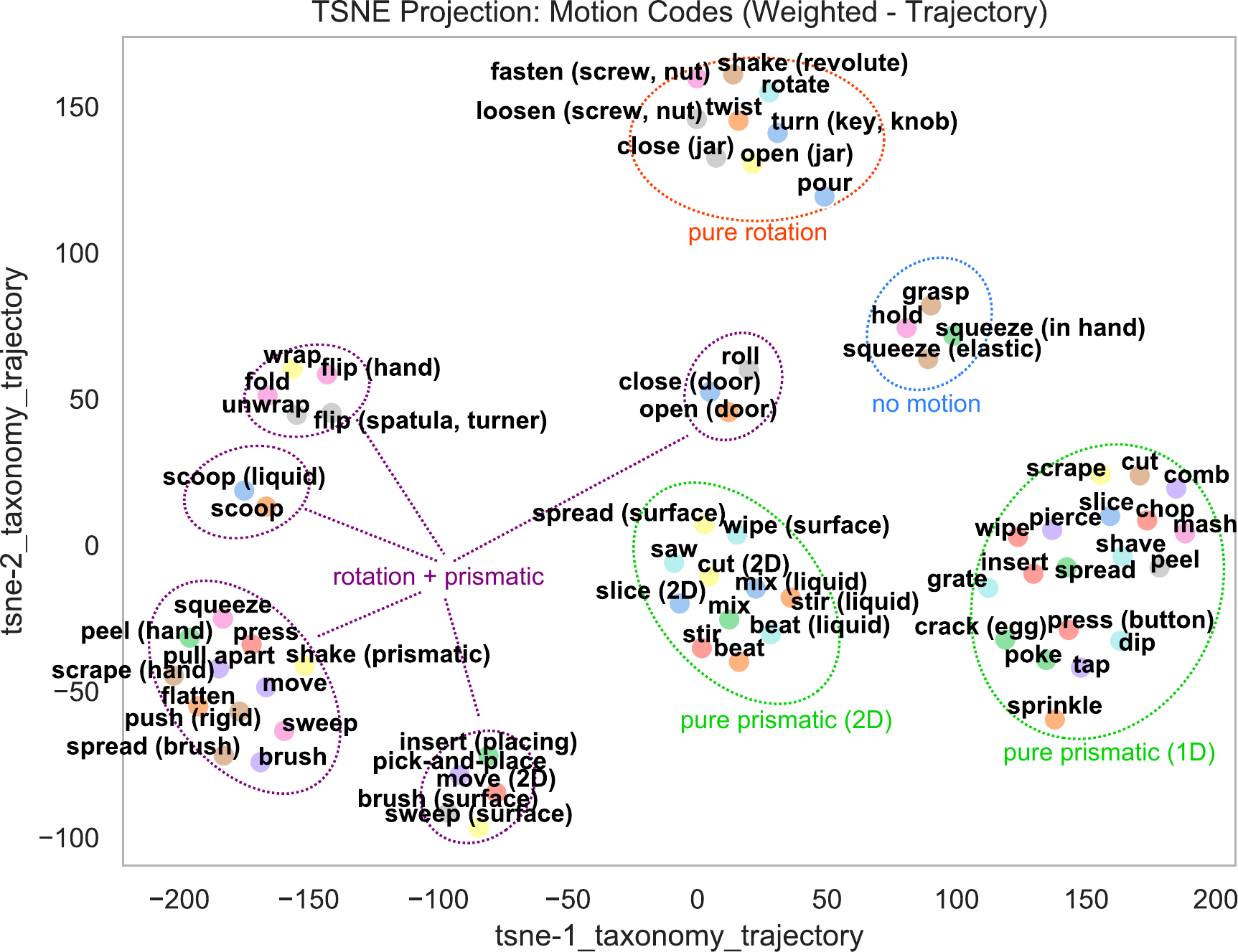}
    \label{fig:embed-mot_t}}
    
    \subfloat[Motion Codes (Hamming distance)]{\includegraphics[width=8.5cm]{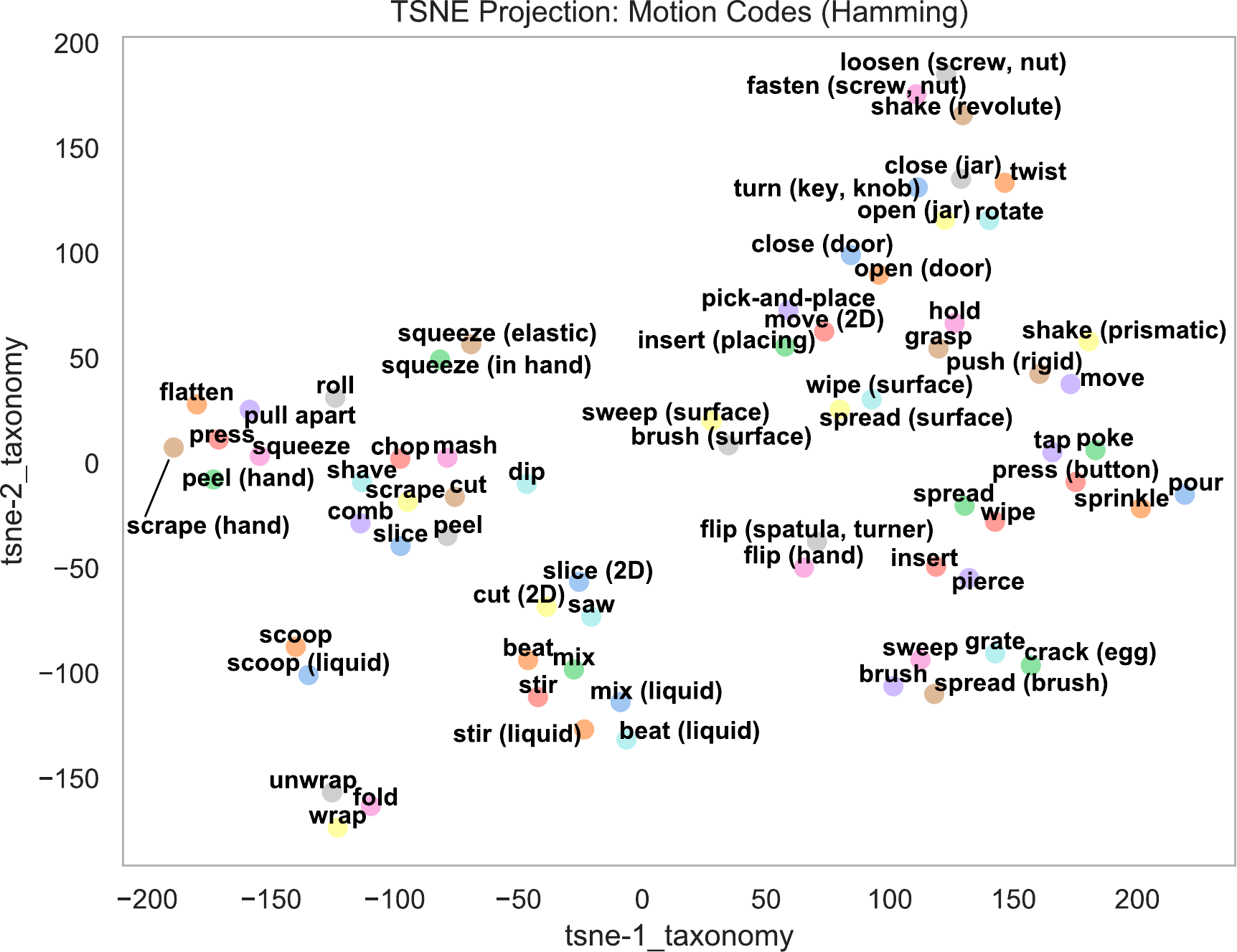}
    \label{fig:embed-mot_ham}}
    \subfloat[Concept-Net~\cite{speer2017conceptnet}]{\includegraphics[width=8.5cm]{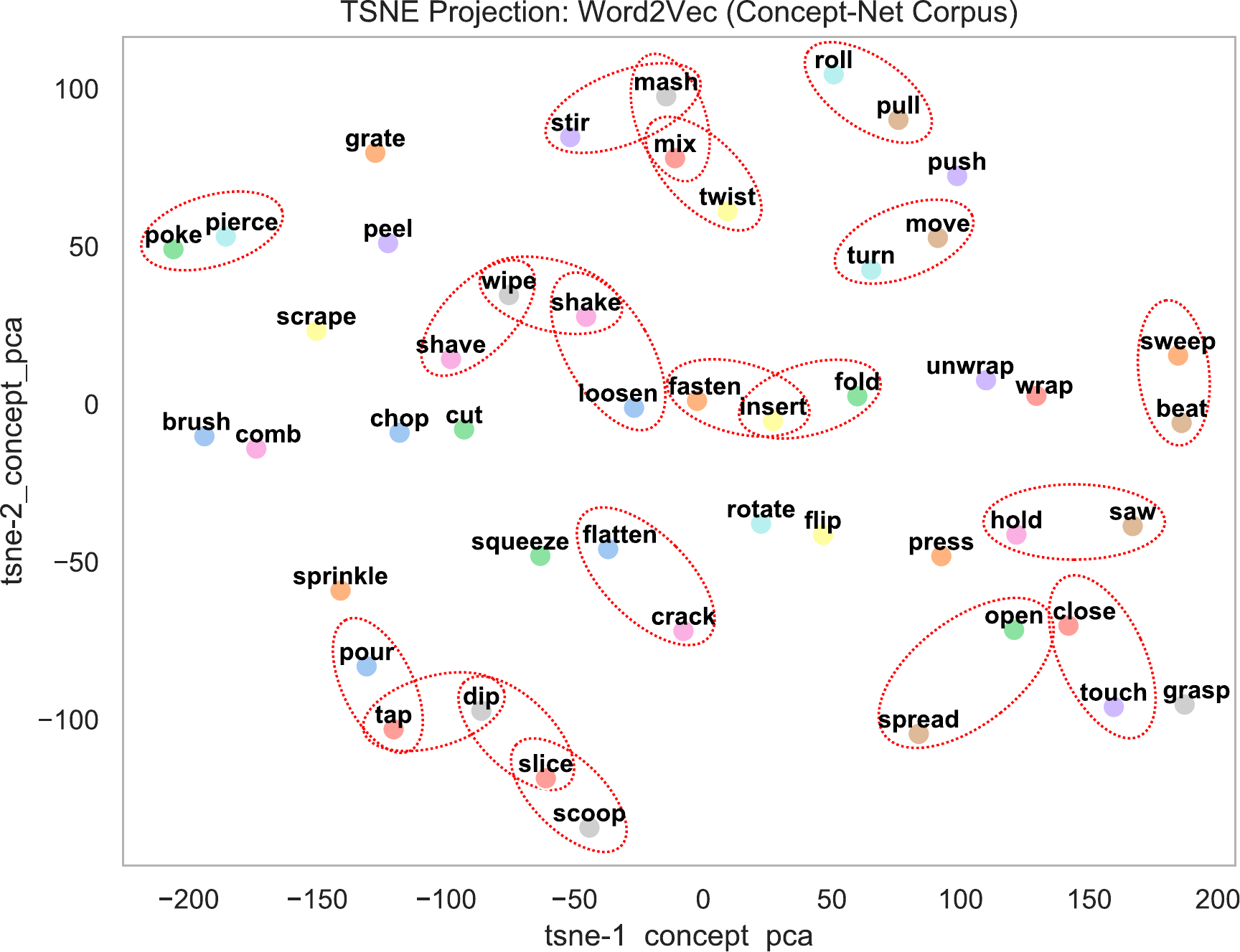}
    \label{fig:embed-con}}

    \subfloat[Google News~\cite{mikolov2013efficient}]{\includegraphics[width=8.5cm]{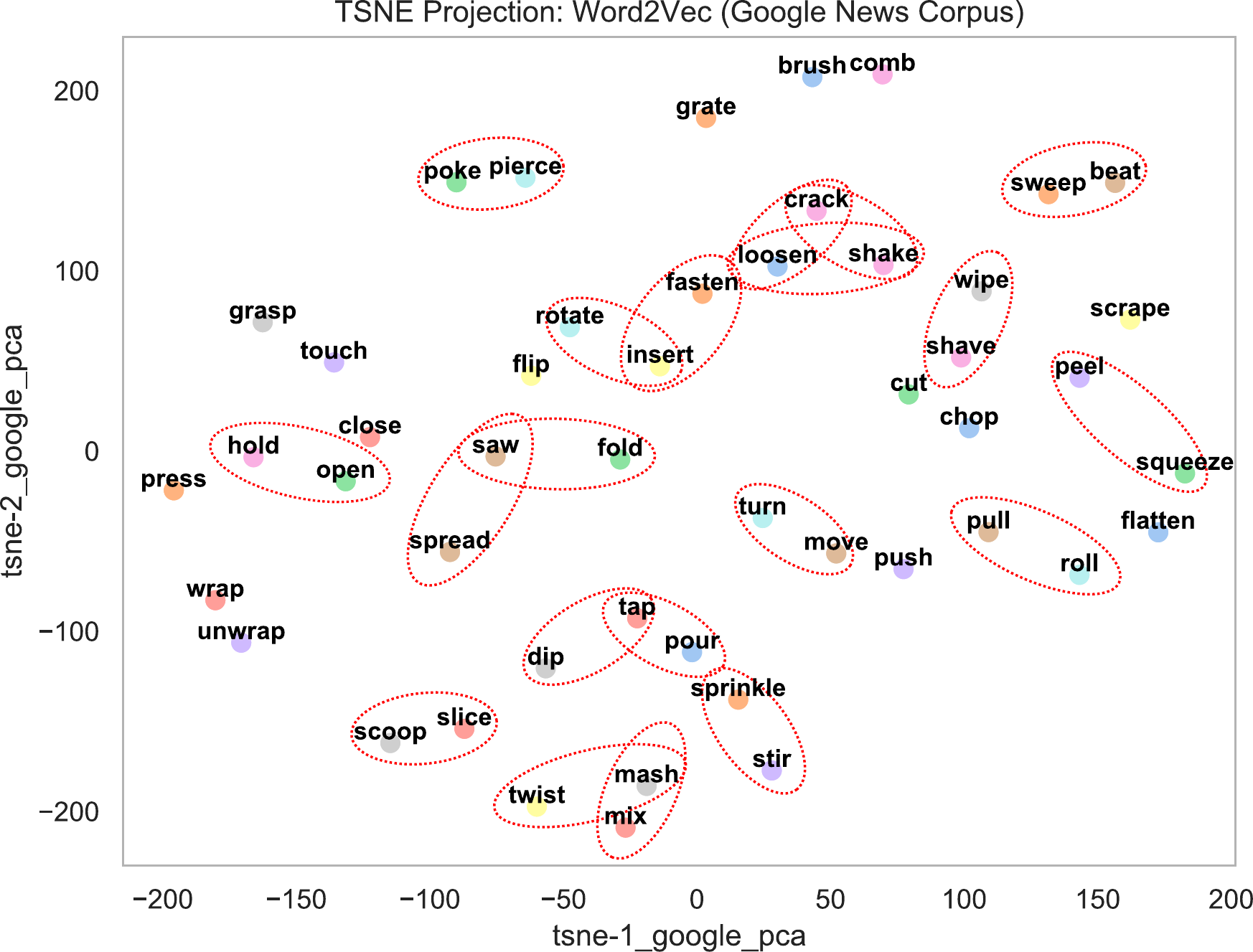}
    \label{fig:embed-goog}}
    \subfloat[Wikipedia 2018~\cite{yamada2018wikipedia2vec}]{\includegraphics[width=8.5cm]{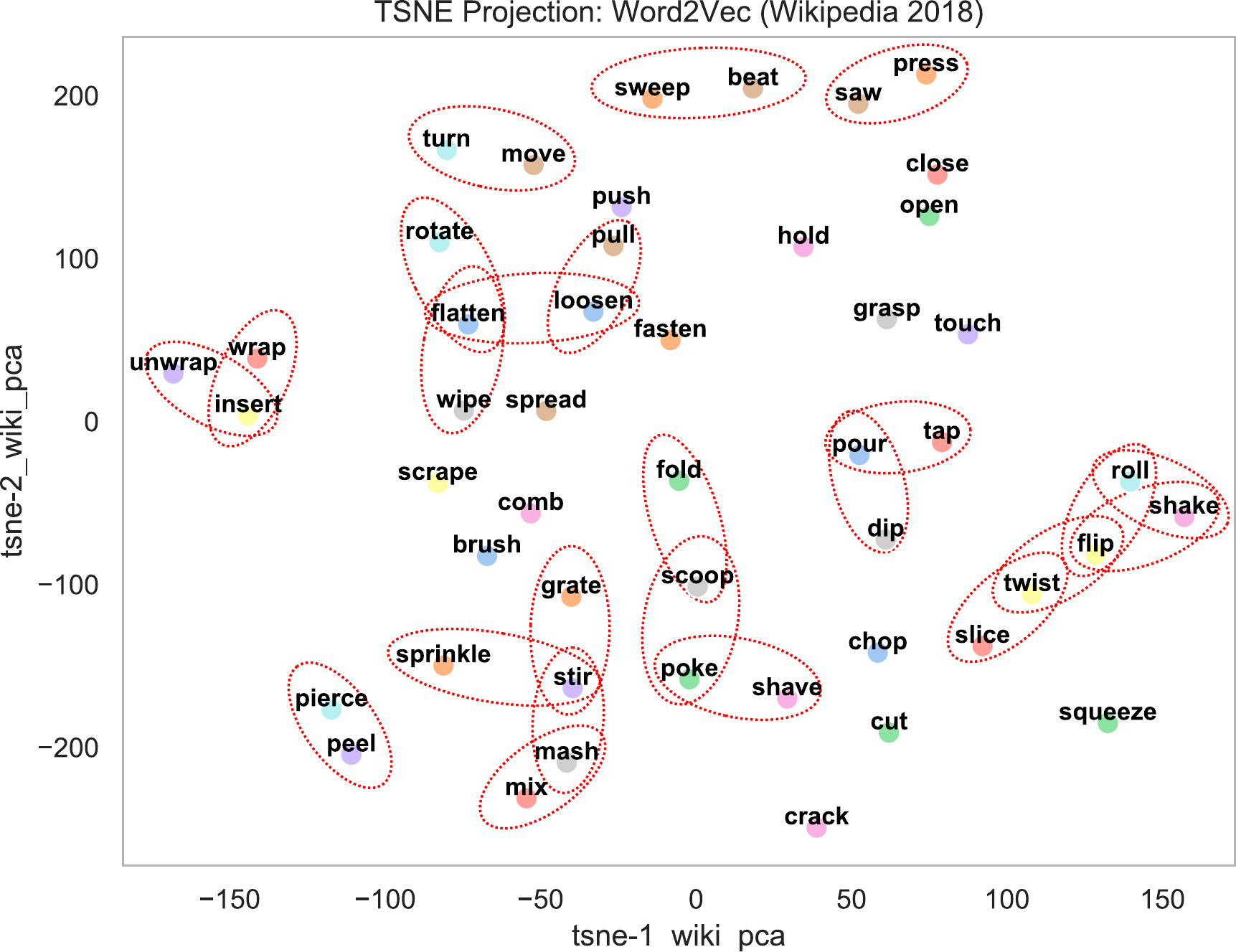}
    \label{fig:embed-wiki}
    }
    \caption{Graphs showing the 2D projection of vectors as a result of t-SNE from: a) motion codes with more weight on contact features, b) motion codes with higher weight on trajectory features, c) motion codes with regular Hamming distance, and Word2Vec embeddings from d) Concept-Net, e) Google News, and f) Wikipedia 2018 (best viewed in colour). 
    We highlight certain examples of motions that do not share mechanical equivalences or similarities in d) - f) in red, and we highlight clusters of similar motions produced by motion codes in a) and b), which use weights to determine distances (in varying colours to distinguish by characteristics).}
    \label{fig:embed}
\end{figure*}

\subsection{Comparing Motion Codes to Word2Vec Embedding}
To show how motion codes produce more ideal distances between motion types, we show how vectors from Word2Vec, which are derived from natural language, are not sufficient to represent manipulations for classification.
As mentioned before, Word2Vec is an unsupervised method to derive multi-dimensional vectors for words in natural language processing tasks with neural networks.
Typically, all words in a vocabulary are initialized as random vectors, whose distances are incrementally adjusted with respect to other word vectors.
Words are related based on locality; that is to say, if one word is frequently seen among neighbours of that word in source text, then its vector along with its neighbouring words' vectors will be closer to one another than those vectors representing other words in the vocabulary.

To compare motion codes to Word2Vec embeddings, we applied dimension reduction with PCA and then used t-SNE \cite{maaten2008visualizing} to visualize these embeddings and their relative distances in 2D.
The t-SNE algorithm (short for \textit{t-distributed Stochastic Neighbor Embedding}) is an approach that is often used to visualize high-dimensional representations or embedding, such as word vectors from Word2Vec, in a low-dimensional space.
Although certain motions will be assigned the same code, the t-SNE algorithm will position their projected vectors in close yet non-overlapping positions; similar motions would be clustered near each other since t-SNE preserves local neighbours while keeping dissimilar motions far from one another.
By default, the distances between the naturally occurring clusters are set further apart than by default, which is reflected by an early exaggeration value of 36 (as opposed to 12); in addition, the number of neighbours used to decide on placement, which is known as perplexity, was set to 12.
Since word vectors from Word2Vec are associated with single words, vectors of functional variants of labels that we have listed in Table \ref{tab:motion_code} cannot be found directly.
For instance, the labels `mix' and `mix (liquid)' are different based on the permanence of deformation.
To circumvent this limitation, some motions were substituted with other words, such as `pick-and-place' to `move', that may capture the meaning of the original label.

In Figures \ref{fig:embed-mot_c}, \ref{fig:embed-mot_t}
and \ref{fig:embed-mot_ham}, we show 2-dimensional projections based on motion codes, while in Figures \ref{fig:embed-con}, \ref{fig:embed-goog} and \ref{fig:embed-wiki}, we see the 2-dimensional projection of motions based on pre-trained Word2Vec models from Concept-Net, Google News and Wikipedia.
Distances in t-SNE for Word2Vec vectors were measured using the cosine similarity metric; with motion codes, we used the regular Hamming metric (Figure \ref{fig:embed-mot_ham}) and a weighted distance metric that we defined ourselves.
Using a weighted approach allows us to emphasize dissimilarity based on key motion taxonomy attributes rather than the regular Hamming metric, which measures the degree of dissimilarity among bits with no considerations for their meanings.

Rather than simply setting the penalty of dissimilarity to 1 for different 
We defined two weighted values, $\alpha$ and $\beta$, which are used to set the priority of contact or trajectory types when measuring distances.
$\alpha$ is a penalty applied when two motions are of different interaction type (i.e. contact versus non-contact),\ as well as contact duration and engagement type, which is reflected by the 1st to 7th most significant bits (MSB);
$\beta$ is a penalty applied for trajectory types, reflected by the 8th to 12th MSB (active trajectory) and 13th to 17th MSB (passive trajectory) of the motion code; specifically, if one motion code exhibits movement and another does not, $\beta$ is added to their distance value, but if they simply differ by the number of axes (1, 2 or 3 DOF), then only half of $\beta$ is added.
All other attributes were measured normally with a penalty of 1.
We illustrate the difference between these distance variations for t-SNE as Figures \ref{fig:embed-mot_c} and \ref{fig:embed-mot_t} respectively.
In Figure \ref{fig:embed-mot_c}, a higher weight is assigned when two motion code vectors are different in interaction type (contact), while Figure \ref{fig:embed-mot_t} places more emphasis on motion trajectory type.
In these figures, we also highlight naturally occurring clusters and neighbouring motion codes that share common attributes.

\subsection{Discussion on Word2Vec Comparison}
As seen in the t-SNE plots in Figure \ref{fig:embed}, using motion codes (from Table \ref{tab:motion_code}) for embedding will result in the placement of functionally similar motions  close to one another (while distancing those that are functionally different as other clusters) in a different way to Word2Vec embeddings.
Using a weighted approach rather than the Hamming distance between motion codes preserves neighbours better.
The major disadvantage of Word2Vec vectors is that we are unable to capture multiple senses or meanings for a single word label.
Furthermore, there is no way of discerning between different forms of a word such as parts of speech. 
For instance, in Figures \ref{fig:embed-goog} to \ref{fig:embed-wiki}, `pour' is placed closest to the word `tap', just as we introduced before.
Since the word `tap' in the English language can either be a verb or noun, the word was interpreted in the context of the noun, as water usually flows or pours out of the tap.
The same can be said of the pair `move' and `turn', which perhaps emphasize the noun meaning more than their verbal meaning.

However, when considering the manipulation in a mechanical sense, it does not match our expectation since their functional attributes are different, where `tap' is considered as contact and prismatic and `pour` is non-contact and revolute.
Instead, using motion codes, if we prioritize trajectory type (Figure \ref{fig:embed-mot_t}), the label `pour' is placed to other revolute-only motions such as `turn (key)', and `fasten (screw, nut)' (although being a cyclical motion); if we prioritize contact interaction type (Figure \ref{fig:embed-mot_c}), the label `pour' was placed closest to the label `sprinkle' since it is also non-contact while being placed further away from contact engagement motions.
Other Word2Vec results that do not match functionality (which we highlight with red ellipses) include `beat'/`sweep', `stir'/`sprinkle' (Figures \ref{fig:embed-goog} and \ref{fig:embed-wiki}), `dip'/`tap', and `mash' and `mix'.
Other than the highlighted motion pairs, Word2Vec embedding generally captured the connection between certain labels such as `cut'/`slice'/`chop' and `sprinkle'/`pour' since these are synonymous to one another.
Another shortcoming of Word2Vec embeddings is that we are unable to compare functional variants of motion types, which was the reason behind simplifying labels to single words.
However, this leads to ambiguity in motion labels since we cannot be very descriptive using one word.
For example, the labels `open door' and `open jar' were simplified as `open', but the sense of opening can differ depending on the manipulated object.
With the two separations `open door' and `open jar', although they serve a similar purpose, the way the motion is executed is different, and these mechanics should be considered when evaluating differences between motions.
Such pairs include `shake' (prismatic and revolute), `mix' (liquid and non-liquid) and `brush' (surface and non-surface).



\section{Conclusion and Future Work}
\label{sec:con}
To conclude, in this paper, based on our work in \cite{paulius2019manipulation}, we proposed an embedding for manipulations better suited for motion recognition in robotics and AI using the \textit{motion taxonomy}.
Embedding with this taxonomy circumvents the issue of language where words can take on multiple meanings. 
One can represent motions using attributes defined in the taxonomy as binary bits, and vectors will describe the mechanics of motions from the robot's point of view.
In our experiments, we demonstrated that these motion codes, when compared to Word2Vec (which uses natural language for training), produce embeddings that provide better metrics for classification.
Furthermore, these features can be extracted directly from demonstration data; with a suitable model, motion codes can be automatically generated.
Motion code features are not limited to those mentioned in this paper, as other attributes could be included that can be extracted directly from data and are more representative depending on the context.
In the future, we will demonstrate how a neural network can automatically generate codes for manipulations in video sequences, after which it would be established whether motion codes improve accuracy in motion recognition tasks.

\section{Acknowledgements}
This material is based upon work supported by the National Science Foundation under Grant Nos. 1812933 and 1910040.

\bibliographystyle{IEEEtran}
\bibliography{ref-IROS19}

\end{document}